%% file: main.tex
\documentclass[runningheads]{llncs}

\usepackage{eccv}

\usepackage{eccvabbrv}

\usepackage{graphicx}
\usepackage{booktabs}
\usepackage{multirow}
\usepackage{pifont}
\usepackage{makecell}
\newcommand{\xmark}{\ding{55}}
\newcommand{\cmark}{\ding{51}}

\usepackage[accsupp]{axessibility}  

\usepackage[colorlinks]{hyperref}
\usepackage[normalem]{ulem}

\usepackage{orcidlink}

\begin{document}

\title{Reprojection Errors as Prompts for Efficient Scene Coordinate Regression} 


\author{Ting-Ru Liu\inst{1} \and
Hsuan-Kung Yang \inst{2} \and
Jou-Min Liu\inst{1}$^*$ \and Chun-Wei Huang\inst{1}$^*$ \and Tsung-Chih Chiang\inst{1}$^*$ \and Quan Kong\inst{2} \and Norimasa Kobori\inst{2} \and Chun-Yi Lee\inst{1}}

\institute{\textsuperscript{1}National Tsing Hua University~~~\textsuperscript{2}Woven by Toyota, Inc. \\
\email{\{tingru, diasimui, alan, george, cymaxwelllee\}@elsa.cs.nthu.edu.tw} \\
\email{\{hsuan-kung.yang, quan.kong, norimasa.kobori\}@woven.toyota} \\
\url{https://tingru0203.github.io/egfs}}

\authorrunning{T.-R. Liu et al.}

\maketitle
\def\thefootnote{*}\footnotetext{indicates equal contribution.}

\input{sections/0_abstract.tex}
\input{sections/1_introduction.tex}
\input{sections/2_related_works.tex}
\input{sections/3_preliminary.tex}
\input{sections/4_motivation.tex}
\input{sections/5_methodology.tex}

\input{sections/6_experiments.tex}
\input{sections/7_conclusion.tex}
\input{sections/acknowledgement.tex}


%
%
\bibliographystyle{unsrt}
\bibliography{egbib}

\end{document}

%% file: sections/0_abstract.tex
\begin{abstract}
Scene coordinate regression (SCR) methods have emerged as a promising area of research due to their potential for accurate visual localization. However, many existing SCR approaches train on samples from all image regions, including dynamic objects and texture-less areas. Utilizing these areas for optimization during training can potentially hamper the overall performance and efficiency of the model. In this study, we first perform an in-depth analysis to validate the adverse impacts of these areas. Drawing inspiration from our analysis, we then introduce an error-guided feature selection (EGFS) mechanism, in tandem with the use of the Segment Anything Model (SAM)~\cite{sam}. This mechanism seeds low reprojection areas as prompts and expands them into error-guided masks, and then utilizes these masks to sample points and filter out problematic areas in an iterative manner. The experiments demonstrate that our method outperforms existing SCR approaches that do not rely on 3D information on the Cambridge Landmarks and Indoor6 datasets.
\end{abstract}

%% file: sections/1_introduction.tex
\section{Introduction}
\label{sec::introduction}
The objective of visual localization is to estimate a 6-DoF camera pose from images, a key component in fields such as Augmented Reality, Virtual Reality, and autonomous driving. Contemporary leading methods in visual localization typically involve establishing 2D-3D correspondences and then utilizing Perspective-n-Point (PnP)~\cite{pnp} with RANSAC~\cite{ransac} for camera pose estimation. These methods can be broadly classified into two main directions: feature-matching~\cite{sarlin2019coarse} and scene coordinate regression (SCR)~\cite{scorf}. Feature-matching approaches reconstruct a 3D scene using Structure from Motion (SfM), identify and describe key points in 2D images~\cite{detone2018superpoint, sun2021loftr}, and link these to 3D coordinates~\cite{sarlin2020superglue, lindenberger2023lightglue}. Nevertheless, they may encounter challenges such as high computational demands, significant storage requirements, and potential privacy concerns~\cite{speciale2019privacy}. On the other hand, SCR methods~\cite{anglerepro, fullframe, dsac, dsacpp, dsacstar, esac, ace} employ deep neural networks (DNNs) to predict the 3D coordinates of pixels and then utilize PnP with RANSAC for camera pose estimation, which provide benefits such as accuracy in smaller scenes, reduced training times, as well as minimized storage requirements. Given these advantages, SCR is thus the primary focus of this study and presents potential for further enhancements.

\begin{figure}[t]
  \centering
  \includegraphics[width=\linewidth]{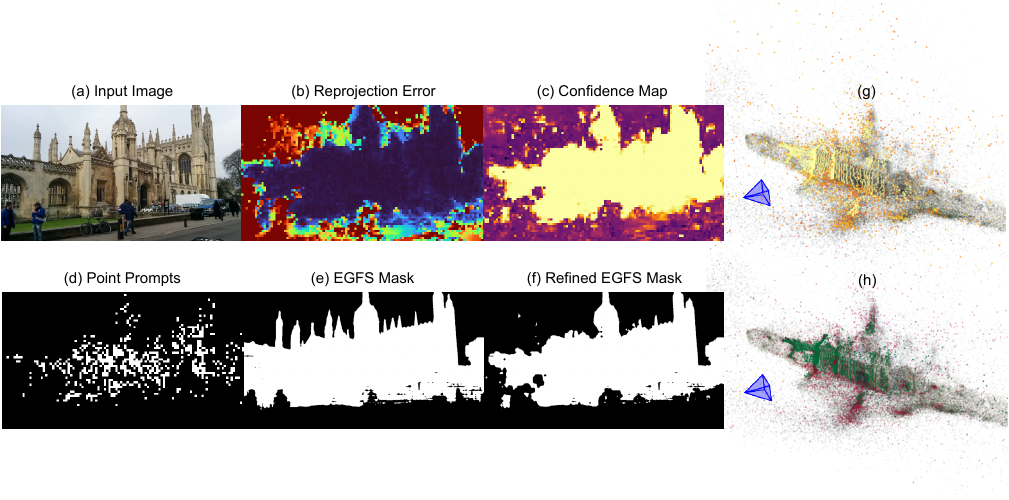}
  \caption{Visualization of the primary components (i.e., (d)-(h)) introduced in the proposed visual localization scheme. (d) illustrates the point prompts selected from (b) with low reprojection errors, while (e) presents an error-guided mask expanded from the prompted points in (d) using SAM. (f) displays the proposed error-guided feature selection (EGFS), which refines the mask from (e) with the predicted confidence map (c) to ensure high-quality scene coordinates are sampled for estimating the final camera pose. The point cloud constructed from the predicted scene coordinates is shown on the right-hand side (i.e., (g)-(h)), with the confidence (yellow parts) and the refined EGFS mask (green for selected areas; red for rejected areas).}
  \label{fig:teaser}
\end{figure}
Albeit effective, previous SCR methods face two primary challenges to be addressed: the presence of dynamic objects and texture-less regions. Dynamic objects, such as pedestrians and cars, pose difficulties for these techniques due to their changing nature and the unsuitable features extracted from these regions. As illustrated in Fig.~\ref{fig:teaser}, these regions could lead to high reprojection errors during training. The second challenge is that current methods struggle with flat, texture-less surfaces. Despite their apparent simplicity, these areas often result in inaccurate scene coordinates due to difficulties in feature extraction. While previous methods~\cite{dsacpp, dsacstar, ace} employ techniques like RANSAC to filter out outliers and differentiable end-to-end optimization approaches to disregard these outliers, they demand significant computational time for end-to-end processing and do not account for the semantic meaning of the selected areas, which may still result in the selection of outlier areas. Thus, they might not effectively prevent the models from unstable weight updates and can lead to training instability.

In light of the aforementioned issues, this study aims to explore the possibility of guiding the sampling process to favor regions with low reprojection errors, and seeks to leverage visual information in an image to expand masks that bear semantic regional meanings. The core philosophy of our methodology focuses on selecting robust regions for training without the need to manually define explicit areas or categories. Such robust regions might differ across scenes, which make it inappropriate for manual definition, as elaborated in Section~\ref{sec::motivation}. More specifically, we propose a strategy, named \textit{Error-Guided Feature Selection (EGFS)}, for deriving low reprojection error samples as point prompts, and expanding them to encompass a complete mask with similar semantic meanings. The masks are iteratively updated after being frozen for a specified period, which enable the proposed model to dynamically renew the focused regions. To achieve this prompt-based semantic regions expansion, we employ the Segment Anything Model (SAM)~\cite{sam}, a vision foundation model capable of providing a general understanding of the scene. This error-guided scheme ensures that our training process concentrates on regions with low reprojection errors and adaptively reduces the variability introduced by dynamic or texture-less objects. Moreover, EGPS adopts a predicted confidence map to refine the expanded error-guided mask for updates, which ensures the regions marked by the masks are sufficiently reliable. An illustrations of each component is provided in Fig.~\ref{fig:teaser}.

To validate the effectiveness of our proposed method, we conduct evaluations on both outdoor and indoor datasets, including the Cambridge Landmarks dataset~\cite{posenet} and the Indoor6~\cite{scenelandmark} dataset, for evaluating the performance of our proposed SCR methodology. The experimental results suggest that our method can be broadly applied to previous methods and provides benefits, and leads to state-of-the-art (SOTA) performance when compared with SCR approaches without leveraging any 3D information in terms of estimated translational and rotational errors, while requiring less training time and smaller model sizes. We further perform a series of detailed analyses and ablation studies to confirm the effectiveness of our error-guided sampled regions and the proposed methodology.

%% file: sections/2_related_works.tex
\section{Related Work}
\label{sec::related_works}

\paragraph{Scene Coordinate Regression.}
Conventional SCR approaches typically begin by establishing 2D-3D correspondences and then utilize RANSAC-based optimization to estimate the camera pose. To establish 2D-3D correspondences, previous endeavors have explored the use of random forests~\cite{scorf, multioutput, forest_rgb, forest_rgbd} or convolutional neural networks (CNNs)~\cite{anglerepro, fullframe, dsac, dsacpp, dsacstar, ace, ofvl} to regress scene coordinates from images. Since SCR typically requires scene-specific training, training time has emerged as a critical concern in practical applications. The ACE~\cite{ace} method, which can be trained within five minutes, stands out as a promising solution. Moreover, unlike methods that map every 2D pixel to a 3D coordinate, some works~\cite{vsnet, scenelandmark, sldstar} have aimed at learning to detect 3D landmarks to establish 2D-3D correspondences. However, they often require the reconstruction of a 3D model, which incurs additional computational time and costs. Our method seeks to utilize SAM to derive error-guided masks from 2D images, while maintaining the training efficiency.

\paragraph{Emphasis on Robust Features for Localization.}
Several prior studies~\cite{pixloc, vsnet, pixselect, global_instance, sfd2, d2s, FocusTune} in visual localization have discovered that not all regions within an image contribute equally to localization accuracy. As a result, these approaches have attempted to distinguish robust and invariant features, found that prioritizing the recognition and comprehension of those regions can enhance visual localization tasks more efficiently and effectively. However, a common limitation of these methods is their reliance on pre-defined semantics or the necessity of a 3D model, which could hinder their generalizability and practicality in real-world applications.

%% file: sections/3_preliminary.tex
\section{Preliminary of Scene Coordinate Regression (SCR)}
\label{sec::preliminary}

SCR-based visual localization determines the camera pose of an RGB image $I \in \mathbb{R}^{H \times W \times 3}$ by predicting 3D scene coordinates $\mathcal{Y}$ for a set of pixels and establishing the 2D-3D correspondences between pixel coordinates in the image and those predicted 3D scene coordinates.
The primary objective of the training process is to establish a mapping function $f(\cdot)$ such that $\mathcal{Y} = f(I)$.  The training data comprise RGB images paired with their respective camera poses, but do not necessarily include depth information. To generate accurate $\mathcal{Y}$, SCR methods often involve optimizing the reprojection error $\mathcal{R}$, which is measured by projecting a 3D predicted scene coordinate $\mathrm{y_i} \in \mathcal{Y}$ back onto its 2D pixel coordinate and calculating the discrepancy from the actual coordinate of the image patch, where $\mathrm{i}$ denotes the patch index. The reprojection error for each patch $\mathrm{r_i} \in \mathcal{R}$ can be formulated as $r(\mathrm{p_i}, \mathrm{y_i}, \mathrm{h^*}) = \|\mathrm{p_i} - K \mathrm{h^{*-1}} \mathrm{y_i}\|$, where $K$ denotes the camera intrinsic matrix and $\mathrm{h^*}$ denotes the ground truth camera pose. After obtaining $\mathcal{Y}$, the camera pose $\mathrm{h}$ can be estimated according to the following equation:
\begin{equation}
\mathrm{h} = g(\mathcal{C}) = g^{PnP}(\mathcal{C_I}), \text{with } \mathcal{C} = \{(\mathrm{p_i}, \mathrm{y_i})|\mathrm{p_i} \in I, \mathrm{y_i} \in \mathcal{Y}\},
\end{equation}
where $g(\cdot)$ involves PnP with RANSAC, followed by Levenberg–Marquardt-based refinement, $\mathcal{C}$ represents the set of all 2D-3D correspondences and $\mathcal{C_I}$ signifies inlier correspondences after refinement, which is a subset of $\mathcal{C}$. Inlier correspondences denote those that are better consistent with the estimated camera pose, while correspondences that do not align well with the estimated camera pose are termed outliers. The final camera pose estimation is determined by the inlier correspondences $\mathcal{C_I}$. In the following paragraphs, we introduce two scene coordinate regression methods that have demonstrated superior performance.
\paragraph{DSAC Variants.}
DSAC~\cite{dsac} introduces a differentiable variant of RANSAC, and enables end-to-end training of the entire pipeline. This design combines the benefits of RANSAC for handling outliers with the power of deep neural networks to learn complex patterns. In addition, DSAC++~\cite{dsacpp} removes the requirement of depth information for SCR training and adopts robust loss terms to downweight outliers. DSAC*~\cite{dsacstar} further simplifies and refines DSAC++, and encourages the network to focus on reliable scene structures while ignoring outlier predictions.

\paragraph{ACE.}
ACE~\cite{ace} is a fast SCR method that can achieve promising localization accuracy within a short training time of only five minutes. ACE splits its regression network into a pre-trained scene-agnostic backbone $f_B$, and a scene-specific multi-layer perceptron (MLP) $f_H$. The backbone $f_B$ extracts feature vectors $\mathrm{f}_i$ from image patches $p_i$ and the MLP $f_H$ predicts the scene coordinates $y_i$ based on the extracted $\mathrm{f}_i$. Between $f_B$ and $f_H$, the features are shuffled to decorrelate gradients within a batch, which enables ACE to enhance its training efficiency.

%% file: sections/4_motivation.tex
\section{In-Depth Evaluation of Scene Coodinate Regression}
\label{sec::motivation}

\subsection{Challenges in Scene Coordinate Regression}
Several challenging factors have been identified that may compromise the accuracy of SCR. Two key challenges among them are dynamic objects and textureless surfaces~\cite{rio10, dong2021robust, Lee_2021_CVPR, pixselect}. Dynamic objects exhibit characteristics of motion or disappearance across different frames, while textureless surfaces lack distinctive feature points or salient characteristics. These challenges highlight the complexity of achieving precise SCR in complex visual scenarios. To address them, previous SCR approaches~\cite{dsac, dsacpp, dsacstar, ace} have employed RANSAC to filter out outliers, and some researches~\cite{dsacpp, dsacstar, ace} have attempted to utilize end-to-end learning to concentrate on reliable scene structures while down-weighting outlier predictions during training. Unfortunately, these prior endeavors failed to fully exploit the rich semantic information in the visual content, which can be crucial for enhancing the robustness and accuracy of camera pose estimation in complex scenes. Furthermore, RANSAC-based approachs for camera pose estimation could potentially fail in scenarios with insufficient number of inlier correspondences~\cite{Fan_2022_CVPR}.

\begin{figure}[t]
  \centering
  \includegraphics[width=.95\linewidth]{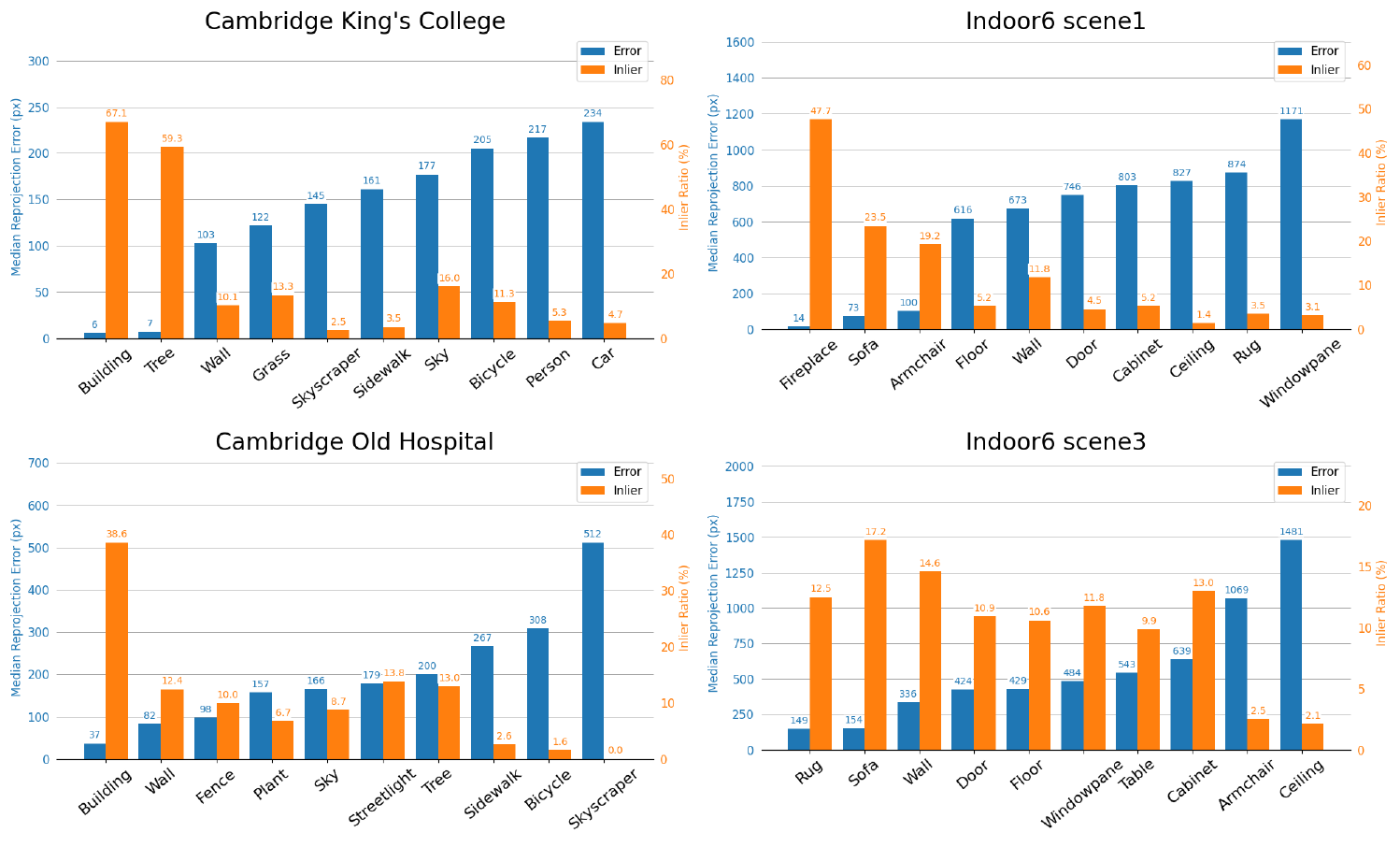}
  \caption{Analysis between reprojection error and semantic meaning. The analysis result indicates the regions with low reprojection errors tend to have higher inlier ratios, while the errors do not always align with specific semantic categories, e.g., ``tree” and ``rug”.}
  \label{fig:ace_analysis}
\end{figure}

\subsection{Analysis between Reprojection Error and Semantic Meaning}
\label{subsec::ace_analysis}
For a more comprehensive understanding of the correlation and relationship between reprojection errors and semantic meaning, we investigate the performance of ACE on the reprojection errors and the inlier correspondences selected by RANSAC for estimating the final camera pose during the inference process. First, image pixels are labeled by a semantic segmentation model ViT-Adapter~\cite{vitadapter}. Subsequently, the reprojection errors and inlier correspondences of ACE for each semantic class are calculated. Fig.~\ref{fig:ace_analysis} presents the median reprojection errors and the ratios of inlier correspondences for the top ten most frequent classes in each scene. The left two figures correspond to the outdoor dataset Cambridge Landmarks~\cite{posenet}, while the right two figures pertain to the indoor dataset Indoor6~\cite{scenelandmark}. It is observed that regions with low reprojection errors tend to have relatively higher inlier ratios, indicating a greater impact on the estimated camera poses. As a result, improving the accuracy of low-error areas could lead to more precise camera pose predictions. Nevertheless, it is also observed that the reprojection errors for the same class can vary significantly across different scenes. For example, in the Cambridge Landmarks dataset, the error for ``tree” is low in King's College but high in Old Hospital. Similarly, in the Indoor6 dataset, the error for ``rug” is high in scene1 but low in scene3. This indicates that low reprojection errors do not always align with specific semantic categories, which presents a challenge in directly using certain specific semantic segmentation labels for area selection. Motivated by these insights, the primary objective of this study is to develop a methodology that does not rely on pre-defined semantic categories to identify low reprojection error areas for enhancing localization accuracy in SCR.

%% file: sections/5_methodology.tex
\section{Methodology}
\label{sec::methodology}

\begin{figure}[t]
  \centering
  \includegraphics[width=\linewidth]{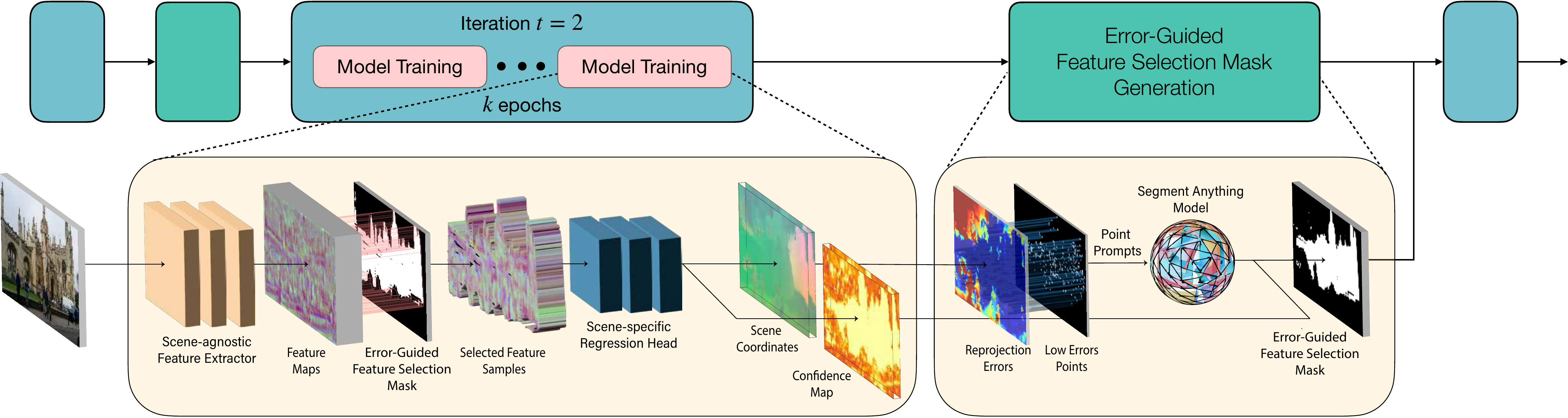}
  \caption{An overview of the training framework.}
  \label{fig:training_overview}
\end{figure}

\subsection{Problem Definition and Framework Overview}
\label{subsec::overview}
\begin{figure}[t]
  \centering
  \includegraphics[width=.7\linewidth]{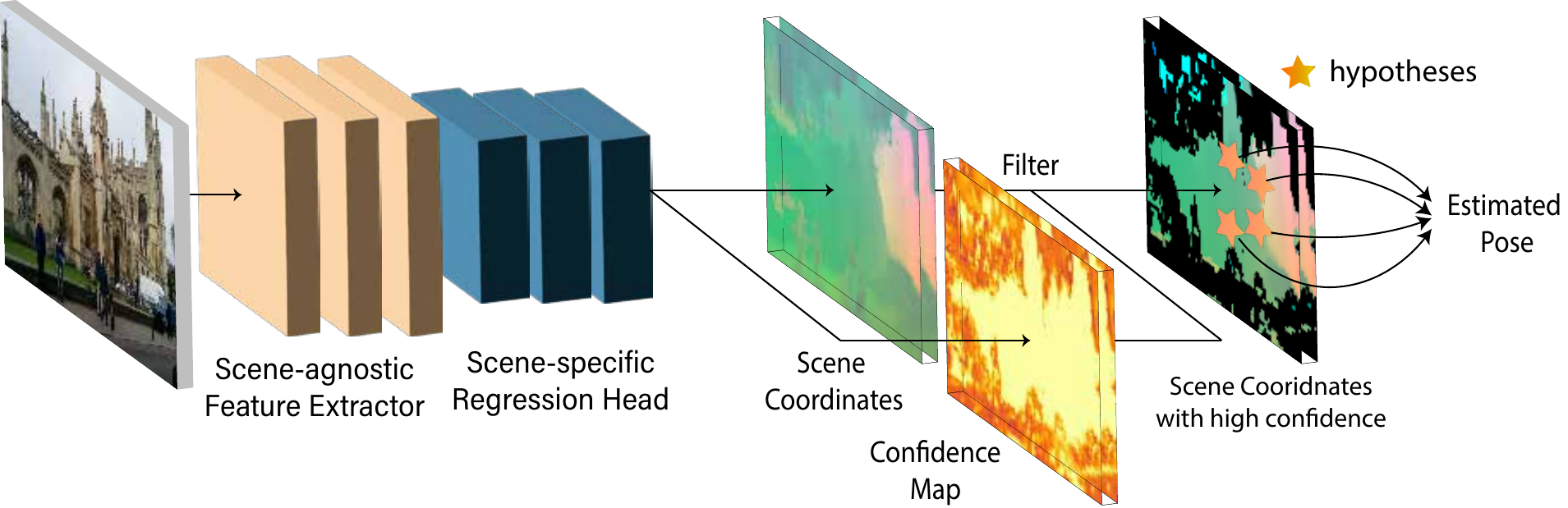}
  \caption{An overview of the inference procedure.}
  \label{fig:inference_overview}
\end{figure}

Fig.~\ref{fig:training_overview} provides an overview of the proposed training framework for SCR, which iteratively samples features to train a scene-specific MLP. The scene-specific MLP consists of a scene coordinate head and a confidence head. In each iteration, the model is trained for $k$ epochs. During the initial iteration, features are randomly sampled from all parts of images in order to derive the first set of reprojection errors. In subsequent iterations, features are selected based on error-guided feature selection (EGFS) masks generated according to reprojection errors and a confidence map. This iterative mechanism enables the model to dynamically update its focus areas throughout the training process. During the inference phase, the network estimates scene coordinates $\mathcal{Y}$ and a confidence map for the refinement, while the masks generated during training are not required for the inference phase. The inference procedure is depicted in Fig.~\ref{fig:inference_overview}.

\subsection{Error-Guided Feature Selection (EGFS) with SAM}
\label{subsec:sam}
In alignment with the discussion in Section~\ref{sec::motivation} on identifying potential areas contributing to the final camera pose estimation during the training stage, we employ the Segmentation Anything Model (SAM)~\cite{sam}, which is capable of leveraging diverse visual cues to generate high-quality object masks. Specifically, in this study, SAM is utilized to extend point prompts with low reprojection errors to form a complete mask sharing a similar semantic context. The rationale behind this approach is the high likelihood that these comprehensive semantic regions correspond to low error areas and can contribute to final pose estimation. Such a concept harnesses the capabilities of the foundational SAM model to uncover more viable areas, and hence, enabling a thorough understanding of the image context. In our experiments, we utilize EfficientViT-SAM-L0~\cite{efficient_vit}, a lightweight and efficient variant of SAM, to identify salient regions by selecting the $\tau$ percentage of points with the lowest reprojection errors as point prompts, where $\tau$ is adjustable and set to ten in our case. After obtaining the error-guided mask from the SAM model, the predicted confidence map $c$ is utilized to refine the error-guided mask to ensure the points marked by the mask is sufficiently reliable. Specifically, in each image, the points with confidence scores below a threshold $\sigma$ are filtered out. The design for confidence refinement and its optimization are described in Section~\ref{sec:conf}. In each iteration, the refined error-guided masks serve as guidance for training the scene-specific MLP. Only the features corresponding to the regions selected by the masks are sampled into the training buffer. This selective inclusion is facilitated by the nature of $f_H$~\cite{ace}, which employs $1\times1$ convolutions to treat and process each selected sample independently using the same set of shared weights. The proposed feature selection mechanism ensures that the training process focuses on the crucial regions identified in the current iteration, and therefore enables the model to enhance its overall performance.

\subsection{Scene Coordinate and EGFS Refinement with Confidence}
\label{sec:conf}
In order to further refine the error-guided mask and ensure the selected points are of sufficiently high quality for predicting accurate camera pose, we incorporate the design of a confidence map~\cite{uncertainty, confnet}. Specifically, we predict a confidence score pixel-wise from the confidence head and optimize it jointly with the re-projection error. This confidence head replaces the last three 1$\times$1 convolution layers of the original scene-specific MLP with two 1$\times$1 convolution layers for confidence score prediction, thus enabling adaptive weights for the reprojection loss at each position. The training loss used by our method is formulated as:
\begin{equation}
\ell(\mathrm{p_i}, \mathrm{y_i}, \mathrm{h_i^*}) =
\begin{cases}
\mathrm{c_i} \cdot \hat{r}(\mathrm{p_i}, \mathrm{y_i}, \mathrm{h_i^*}) - \alpha \log \mathrm{c_i} & \text{if } \mathrm{y_i} \in \mathcal{V} \\
\|\mathrm{y_i} - \bar{\mathrm{y_i}}\|_0 - \alpha \log(1 - \mathrm{c_i}) & \text{otherwise}
\end{cases},
\end{equation}
where $\hat{r}(\mathrm{p_i}, \mathrm{y_i}, \mathrm{h_i^*})$ represents a $\texttt{tanh}$ clamping of the reprojection error that changes over time, 
$\mathcal{V}$ denotes the set of 2D pixels with valid scene coordinate predictions, as explained in~\cite{ace, dsacstar}, $\mathrm{c_i}$ is the confidence score at pixel $\mathrm{i}$, $\alpha$ is a hyperparameter that balances the confidence regularization term, and $\bar{\mathrm{y_i}}$ is a dummy scene coordinate derived from the ground truth camera pose, assuming a constant image depth of 10m. When the scene coordinate prediction is valid, the model down-weights the reprojection errors in challenging regions using confidence scores and focuses more on regions with reliable predictions. On the other hand, when the scene coordinate prediction is invalid, the model is incentivized to lower the confidence scores for inaccurate predictions. The confidence score is also used in the inference phase to select reliable scene coordinates. For each query image, the median of the confidence scores is calculated, and only scene coordinates with confidence greater than this median are selected. This thresholding ensures that only the most reliable 2D-3D correspondences are used to generate the camera pose by the PnP with RANSAC pose solver. By focusing on high-confidence correspondences, a more accurate camera pose can be estimated.

%% file: sections/6_experiments.tex
 \section{Experimental Results}
\label{sec::experiments}

\subsection{Experimental Setups}
\subsubsection{Datasets.} Our evaluation of the proposed method was conducted on two representative datasets: the outdoor dataset Cambridge Landmarks~\cite{posenet} and the indoor dataset Indoor6~\cite{scenelandmark}. The details of the datasets are described as follows:

\input{tables/cambridge.tex}
\input{tables/indoor6.tex}
\input{tables/error_mask.tex}

\paragraph{Cambridge Landmarks.} 
The Cambridge Landmarks Dataset~\cite{posenet} is a renowned outdoor dataset extensively utilized for visual localization tasks, which includes five distinct scenes. Every scene in this dataset is composed of RGB images along with their respective ground truth camera poses reconstructed using the SfM technique. It includes a broad spectrum of conditions and various viewpoints, and is ideal for evaluating the resilience and effectiveness of various SCR methods.

\paragraph{Indoor6.} 
The Indoor6~\cite{scenelandmark} dataset comprises six distinct indoor scenes captured over several days, with ground truth camera poses computed using COLMAP~\cite{sfm}. Each scene includes multiple rooms and contains illumination variations, making it challenging for different types of visual localization tasks.

\subsubsection{Implementation Details.}
The training process of our methodology includes 20 epochs, with the generation of masks every five epochs. We set the confidence regularization parameter $\alpha$ to ten and the reprojection error threshold $\tau$ to ten. For mask refinement, the parameter $\sigma$ is set to the median confidence score of each image. Our method builds upon the ACE~\cite{ace} architecture as the backbone, and thus, the remaining hyperparameters are kept the same as those used in~\cite{ace}.

\subsection{Results on the Cambridge Landmarks and Indoor6 Dataset}
\subsubsection{Cambridge Landmarks Dataset.}
Table~\ref{table:cambridge} presents the experimental results evaluated on the Cambridge Landmarks dataset, which affirm the effectiveness and efficiency of our proposed methodology. It can be observed that EGFS maintains a model size similar to ACE, with only a 0.5MB increase for the confidence head, while reducing average translational and rotational errors. The results further reveal that EGFS not only surpasses DSAC* in performance and performs on par with the ensemble version of ACE (i.e., ACE (\textit{quad.})), but also features a smaller model size and reduced training time. Furthermore, our model's ensemble version (i.e., EGFS (\textit{dual.})) is able to exceed the performance of DSAC* and ACE (\textit{quad.}) while requiring a smaller model size and less training time.

\subsubsection{Indoor6 Dataset.}
Table~\ref{table:indoor6} presents the evaluation results on Indoor6. EGFS significantly outperforms DSAC* and ACE across all scenes, while also achieving comparable performance to ACE (\textit{quad.}) with a much smaller model size (4.5MB compared to 16MB). Moreover, EGFS (\textit{dual.}) demonstrates superior performance relative to ACE (\textit{quad.}), which further substantiates the effectiveness of EGFS in optimization through error-guided masks and confidence refinement.

\subsection{Effectiveness of Error-Guided Feature Selection}
To further substantiate the advantages of employing the EGFS approach, this experiment compares several baseline schemes that sample points with different reprojection error thresholds against EGFS. The results are summarized in Table~\ref{table:error_mask}. Specifically, these baseline schemes select sample points based on various quantiles of the reprojection error map, ranging from 30\% to 70\%, denoted as $Q_{0.3}(R)$ through $Q_{0.7}(R)$, respectively. In each baseline scheme, a point is chosen if its reprojection error $r$ falls below the threshold specified by $Q_{quantile}(R)$. The results suggest that relying solely on reprojection errors does not guarantee optimal learning of scene coordinates. This is due to the fact that areas of low reprojection errors can be scattered and may not encompass entire semantic objects. In contrast, the proposed EGFS expands the points into complete semantic masks followed by confidence refinement, which enables our model to determine scene coordinates with minimal transitional and rotational errors. This experiment thus confirms the effectiveness of leveraging SAM's ability to interpret image context alongside confidence maps for enhanced sample point selection.

\begin{figure}[t]
  \centering
  \includegraphics[width=.9\linewidth]{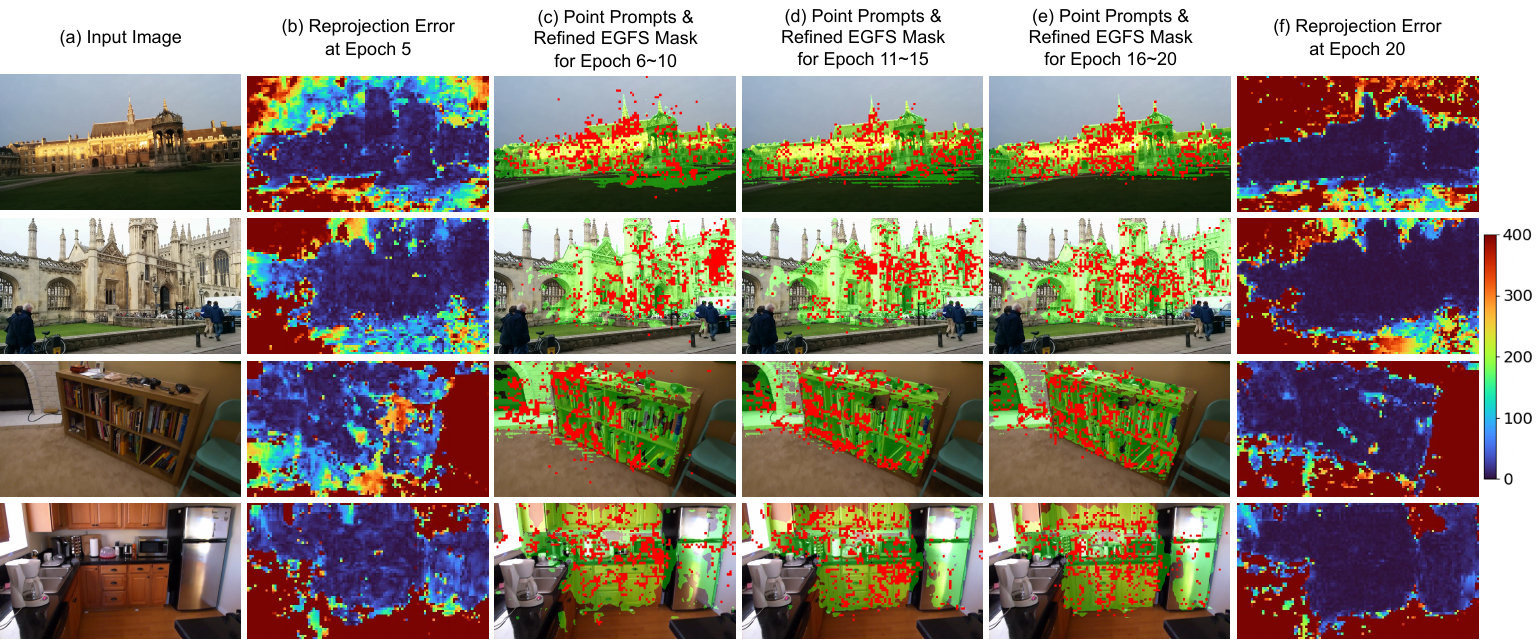}
  \caption{Visualization of the EGFS mask refinement process at every five epochs, which depicts the reprojection errors at the beginning (epoch 5) and the end (epoch 20), as well as the refined error-guided masks used throughout training. The red dots represent low reprojection errors that serve as prompts, while the light green overlay denotes the refined EGFS masks. It can be observed that the EGFS masks enhances over epochs.}
  \label{fig:qualitative}
\end{figure}

\begin{figure}[t]
  \centering
  \includegraphics[width=.9\linewidth]{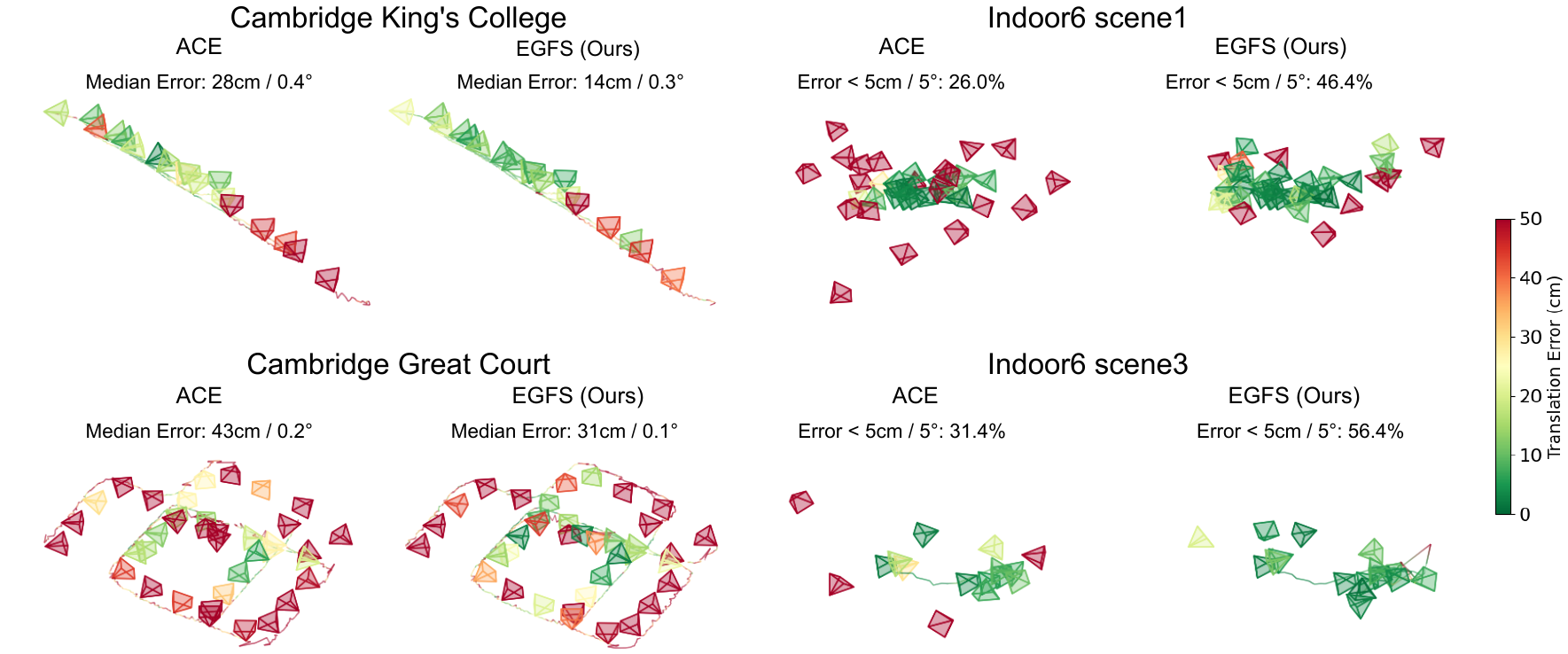}
  \caption{Visualization of the estimated camera pose trajectories from testing sequences with the camera frustums colored based on translational errors. Pose errors denoted.}
  \label{fig:localization}
\end{figure}

\begin{figure}[t]
  \centering
  \includegraphics[width=.88\linewidth]{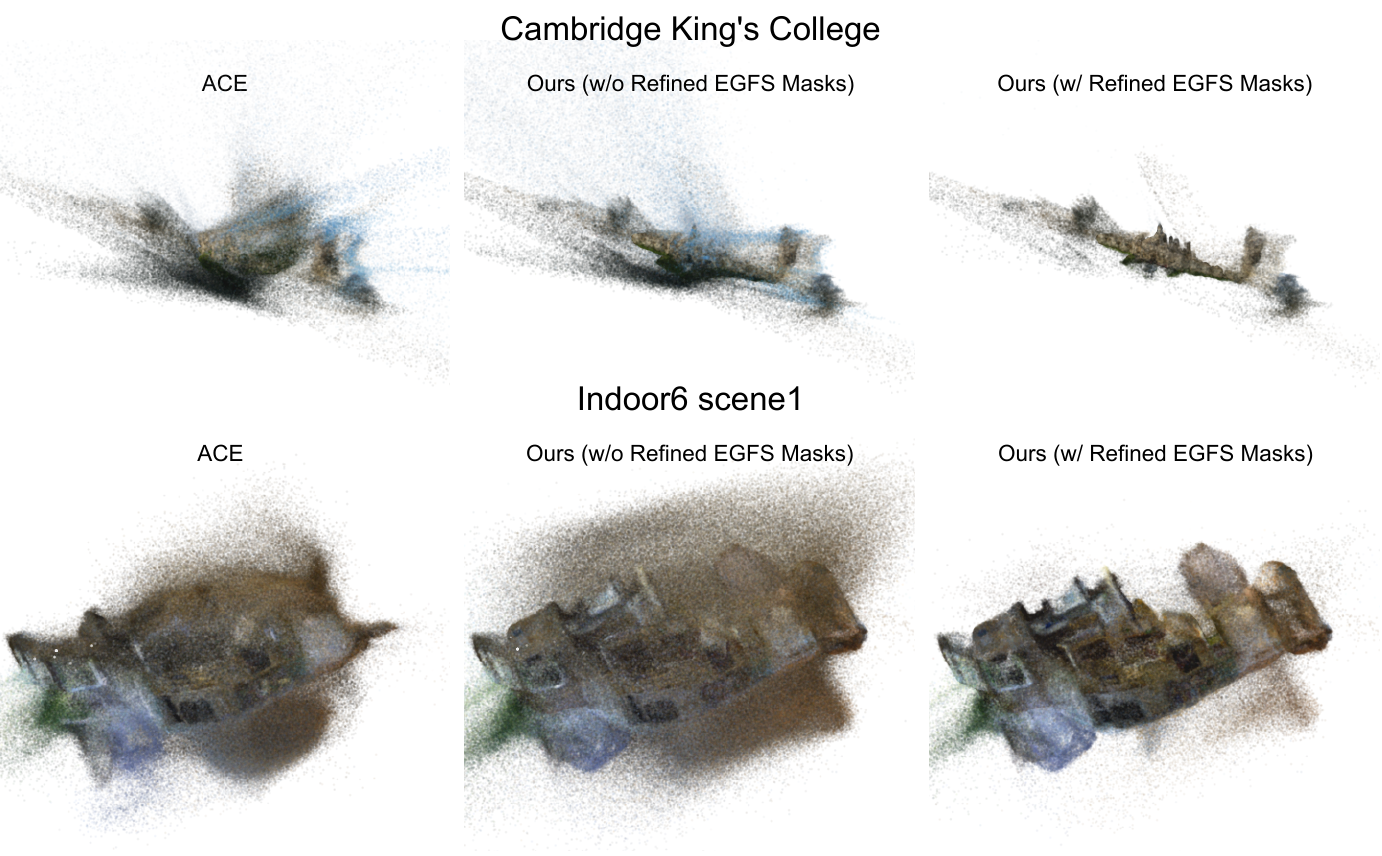}
  \caption{Visualization of point clouds reconstructed from estimated scene coordinates of the training sequence. The point clouds derived from the scene coordinates estimated by the model trained with the proposed EGFS are clearer compared to those from the ACE model, even without the application of refined EGFS masks at inference time.}
  \label{fig:scene_coordinate}
\end{figure}

\subsection{Qualitative Results}
\subsubsection{Iterative Error-guided Feature Selection Mask.}
Fig.~\ref{fig:qualitative} illustrates the refinement of EGFS masks every five epochs and compares the reprojection errors at the beginning and the end. It can be observed that EGFS is not only capable of expanding points with low reprojection errors (i.e., $\mathrm{r} < Q_{0.1}(R)$) into regions that bear similar semantic meanings, but also refines the masks with a confidence map to exclude uncertain areas. The masks are observed to be iteratively enhanced from coarse to fine. In the beginning (i.e., at epoch five), the reprojection errors appear noisy, whereas by the end (i.e., at epoch 20), the EGFS masks become more refined and concentrated on key regions beneficial for SCR.

\subsubsection{Visualization of the Estimated Camera Poses.}
In this section, we present the estimated camera poses from the testing sequences, and compare the proposed EGFS with the ACE baseline. The results are illustrated in Fig.~\ref{fig:localization}. To visualize camera trajectories, we connect consecutive camera positions with lines and indicate camera orientations with frustums. Each sampled camera pose is associated with a color to represent the translation error. It is observed that our proposed EGFS can estimate more accurate camera poses as compared to ACE.

\subsubsection{Scene Coordinates and 3D Point Clouds w/ and w/o EFGS.}
To validate the effectiveness of EGFS qualitatively and its impact on the quality of estimated scene coordinates that learned from training sequence, we depict the 3D point cloud generated from the scene coordinates from the training sequence, and apply colors based on the corresponding queried pixels in Fig.~\ref{fig:scene_coordinate}. To compare the 3D point clouds reconstructed from the scene coordinates, we compare the baseline ACE approach with the proposed EGFS approach. For our approach, we visualize the point clouds both before and after the application of EGFS masks and confidence maps when collecting the scene coordinates. Please note that both before and after versions which shown in the figure are trained with EGFS refinement. The 3D point cloud constructed by the proposed approach appears clearer, in contrast to the ACE-generated point cloud, which exhibits blurriness due to inaccurate scene coordinate estimation. In addition, the point cloud reconstructed from raw scene coordinates predicted without EGFS from our approach contains noise and floaters due to the inclusion of scene coordinates with low confidence scores. On the other hand, the point cloud reconstructed from the scene coordiantes with EGFS refinement exhibits clearer contours and less noise. This justifies EGFS's ability to eliminate areas of low confidence values and produce high-quality dense scene coordinates.

\subsubsection{Visualization of Estimated Confidence Map on Unseen Samples.}
\begin{figure}[t]
  \centering
  \includegraphics[width=.88\linewidth]{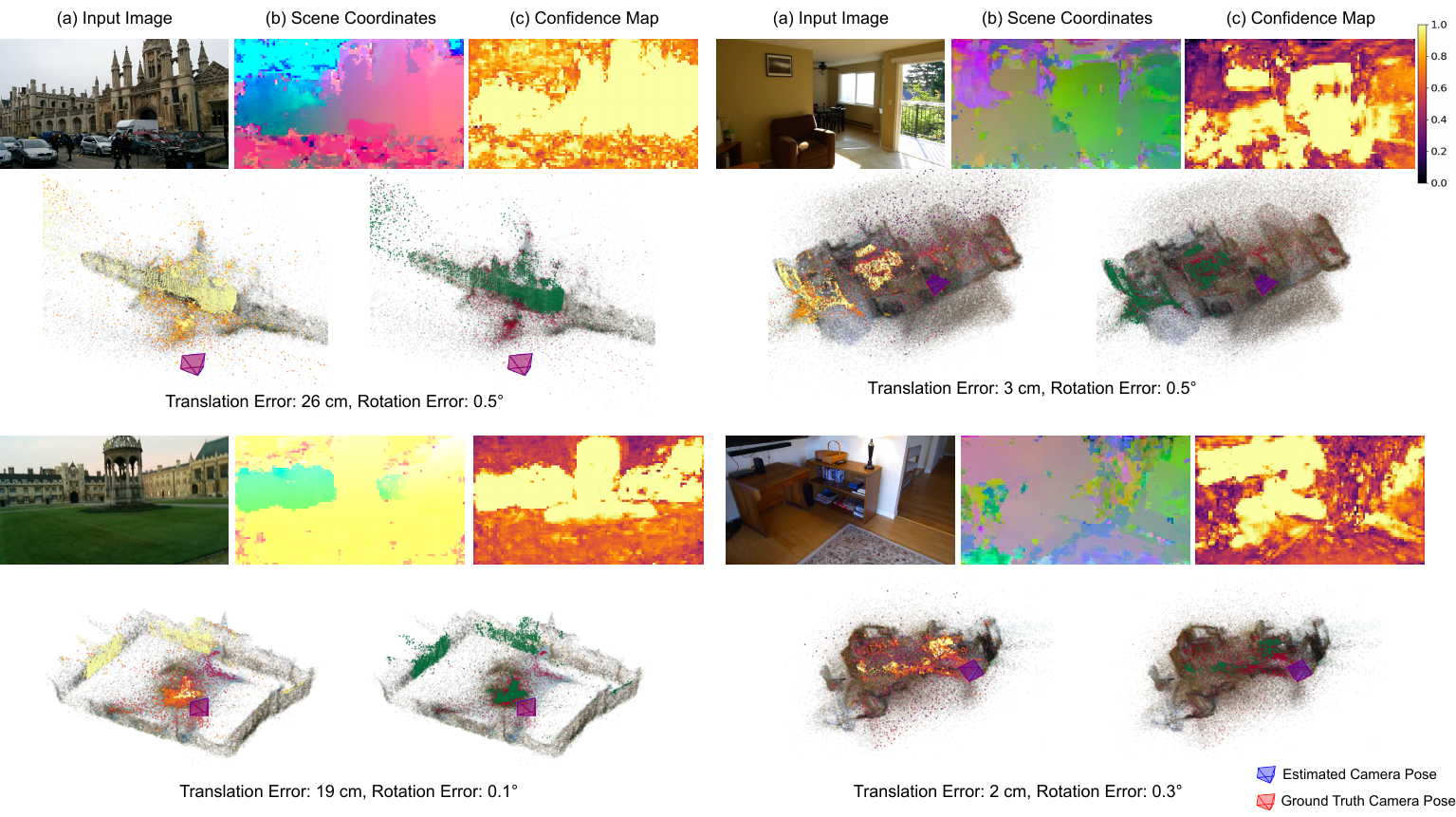}
  \caption{Visualization of the estimated scene coordinates and the confidence maps on unseen samples. The RGB color in the scene coordinates represents the XYZ coordinates. The point clouds are colored green for selected areas and red for rejected areas.}
  \label{fig:confidence}
\end{figure}

In this section, we qualitatively evaluate the confidence maps estimated from the testing sequence by visualizing them in both the 2D image plane and as 3D projections onto the point cloud. The visualized point cloud is reconstructed from the training sequence, whereas the plotted samples are selected from the testing sequence and were not included in the training procedure. These visualizations are provided in Fig.~\ref{fig:confidence}. It can be observed that the estimated confidence map assists in rejecting scene coordinates located in areas where accurate prediction is challenging, such as in the air or on texture-less surfaces.

\subsection{Ablation Study}
\subsubsection{Error-Guided Feature Selection (EGFS) and Confidence Refinement.}
We first present an ablation analysis to validate the effectiveness of the EGFS masks and confidence refinement in enhancing our method's performance. Table~\ref{table:ablation} reports the average results on the Cambridge Landmarks and Indoor6 datasets. The results demonstrate that the model trained with the proposed EGFS mechanism outperforms those trained without it, which indicates that focusing on training with robust features leads to improved performance. Furthermore, the incorporation of the confidence map also enhances the overall performance. The final row of Table~\ref{table:ablation} demonstrates that combining both techniques (i.e., our EGFS) achieves the best results,  which further confirms the effectiveness of EGFS in improving localization accuracy.

\input{tables/ablation.tex}

\subsubsection{Analysis of Proportions of Point Prompts.}
We evaluate the performance of different proportions of point prompts used for expanding into EGFS masks, and the results are reported in Table~\ref{table:error-prompts}. It can be observed that the proportions of point prompts do not significantly impact the performance on the Cambridge Landmark dataset. This may be attributed to the fact that the scenes in the dataset typically feature a primary architectural structure with fewer complicated details in the surroundings. On the other hand, for the Indoor6 dataset, the proportions of point prompts significantly affect the performance. The rationale behind these observations is that the indoor scenes are more complicated, and feature various small items that necessitate a larger number of low-error point prompts to comprehensively capture the entire scene structure. Please note that we employ $\tau=10$ as our setting for all the experiments presented.

%% file: tables/cambridge.tex
\begin{table}[t]
\caption{A comparison of model sizes, training times, and median translation and rotation errors (in cm/$^\circ$) evaluated on the Cambridge Landmarks dataset. `\textit{dual}' and `\textit{quad.}' denote the ensemble versions of two and four models, respectively.}

\centering
\footnotesize
\resizebox{\textwidth}{!}{

\begin{tabular}{clcc|ccccc|c}
\toprule
& Method & Size & Mapping Time & King’s College & Great Court & Old Hospital & Shop Facade & St Mary’s Church &  Average \\ 
\midrule
\multirow{2}{*}{FM} & hLoc~(SP+SG)~\cite{sarlin2019coarse, detone2018superpoint, sarlin2020superglue}  & $\sim$800MB & \multirow{2}{*}{$\sim$35 min} & 12/0.2 & 16/0.1 & 15/0.3 & 4/0.2 & 7/0.2 & 11/0.2\\
                    & pixLoc~\cite{pixloc}  & $\sim$600MB & & 30/0.1 & 14/0.2 &  16/0.3 & 5/0.2 & 10/0.3 & 15/0.2 \\ 
\midrule 
\multirow{5}{*}{SCR} & DSAC*~\cite{dsacstar} & 28MB & 15 hr & 15/0.3 & 49/0.3 & 21/0.4 & 5/0.3 & 13/0.4 & 21/0.3 \\
                    & FocusTune~\cite{FocusTune}        &  4MB & 6 min & 19/0.3 & 38/0.1 & 18/0.4 & 6/0.3 & 15/0.5 & 19/0.3 \\
                     & FocusTune (\textit{quad.})~\cite{FocusTune} & 16MB & 24 min & 15/0.3 & 29/0.1 & 17/0.4 & 5/0.2 & 9/0.3 & 15/0.3 \\
 (w/ 3D model) & NeuMap~\cite{neumap}   & - & - & 14/0.2 &  6/0.1 & 19/0.3 & 6/0.2 & 17/0.5 & 12/0.3 \\
                    & SACReg-L~\cite{sacreg} & - & - & 11/0.2 & 13/0.1 & 13/0.2 & 6/0.2 & 5/0.3 & 10/0.2 \\
\midrule 
\multirow{6}{*}{SCR} & DSAC*~\cite{dsacstar} & 28MB & 15 hr & \underline{18/0.3} & 34/0.2 & \underline{21/0.4} & \underline{5/0.3} & 15/0.6 & 19/0.4 \\
                    & ACE~\cite{ace}        &  4MB & 5 min & 28/0.4 & 43/0.2 & 31/0.6 & \underline{5/0.3} & 18/0.6 & 25/0.4 \\
                    & ACE (\textit{quad.})~\cite{ace} & 16MB & 20 min & \underline{18/0.3} & \textbf{28/0.1} & 25/0.5 & \underline{5/0.3} & \textbf{9/0.3} & \underline{17/0.3} \\
                    \cmidrule(lr){2-10} \morecmidrules \cmidrule(lr){2-10}
                    & EGFS &  4.5MB & 12 min & \textbf{14/0.3} & \underline{31/0.1} & \underline{21/0.4} & \underline{5/0.3} & 15/0.5 & \underline{17/0.3} \\
                    & EGFS (\textit{dual})  & 9MB & 21 min & \textbf{14/0.3} & \textbf{28/0.1} & \textbf{19/0.4} & \textbf{5/0.2} & \underline{10/0.3} & \textbf{15/0.3} \\
\bottomrule

\end{tabular}}
\label{table:cambridge}
\end{table}

%% file: tables/indoor6.tex
\begin{table*}[t]
\caption{A comparison of model sizes, training times, and the proportions of translation and rotation errors that are below 5cm/5$^\circ$ evaluated on the Indoor6 dataset.
}

\centering
\footnotesize
\resizebox{\textwidth}{!}{

\begin{tabular}{clcc|cccccc|c}
\toprule
& Method & Size & Mapping Time & scene1 & scene2a & scene3 & scene4a & scene5 & scene6 & Average \\ 
\midrule
\multirow{1}{*}{FM} & hLoc~\cite{sarlin2019coarse}  & $\sim$1.5GB & $\sim$3.3 hr & 70.5\% & 52.1\% & 86.0\% & 75.3\% & 58.0\% & 86.7\% & 71.4\% \\
\midrule 
\multirow{3}{*}{SCR} & DSAC*~\cite{dsacstar} & 28MB & 15 hr & 18.7\% & 28.0\% & 19.7\% & 60.8\% & 10.6\% & 44.3\% & 30.4\% \\
    & SLD* (300 landmarks)~\cite{sldstar} & 15MB & $\sim$5.5 hr & 47.2\% & 48.2\% & 56.2\% & 67.7\% & 33.7\% & 52.0\% & 50.8\% \\
(w/ 3D model)  & SLD* (1000 landmarks)~\cite{sldstar}        &  120MB & $\sim$44 hr & 68.5\% & 62.6\% & 76.2\% & 77.2\% & 57.8\% & 78.0\% & 70.1\% \\
\midrule 

\multirow{5}{*}{SCR} & DSAC*~\cite{dsacstar} & 28MB & 15 hr & 23.0\% & 33.9\% & 26.0\% & 67.1\% & 10.6\% & 50.2\% & 35.1\% \\
                    & ACE~\cite{ace}        &  4MB & 5 min & 26.0\% & 32.3\% & 31.4\% & 62.0\% & 14.2\% & 47.4\% & 35.6\% \\
                    & ACE (\textit{quad.})~\cite{ace} &  16MB & 20 min & \underline{52.9\%} & 52.5\% & \underline{62.9\%} & 69.6\% & \textbf{31.1\%} & \textbf{82.4\%} & \underline{58.6\%} \\
            
                    \cmidrule(lr){2-11} \morecmidrules \cmidrule(lr){2-11}
                    
                    & EGFS &  4.5MB & 21 min & 46.4\% & \textbf{60.6\%} & 56.4\% & \textbf{78.7\%} & 22.8\% & 71.6\% & 56.1\% \\
                    & EGFS (\textit{dual}) &  9MB & 30 min & \textbf{58.5\%} & \underline{59.1\%} & \textbf{67.0\%} & \underline{76.1\%} & \underline{30.6\%} & \underline{75.9\%} & \textbf{61.2\%} \\

\bottomrule
\end{tabular}}
\label{table:indoor6}
\end{table*}

%% file: tables/error_mask.tex
\begin{table}[t]
\caption{Comparison of the effectiveness between (a) directly sampling solely based on pixel points with low reprojection errors, and (b) sampling through the EGFS masks.}

\centering
\footnotesize

\resizebox{.95\linewidth}{!}{
\begin{tabular}{c | r|ccccc|c}
\toprule
Dataset & Scene & $\mathrm{r}_i<Q_{0.3}(R)$  & $\mathrm{r}_i<Q_{0.4}(R)$ & $\mathrm{r}_i<Q_{0.5}(R)$ & $\mathrm{r}_i<Q_{0.6}(R)$ & $\mathrm{r}_i<Q_{0.7}(R)$ & \textit{EGFS Masks} \\
\midrule
\multirow{5}{*}{\textit{Cambridge}~} & King's College~ & 20/0.3 & 19/0.3 & 20/0.3 & 20/0.3 & 20/0.3 & \textbf{14/0.3} \\
& Great Court~ & 39/0.2 & 37/0.1 & 36/0.2 & 38/0.2 & 39/0.2 & \textbf{31/0.1} \\
& Old Hospital~ & 26/0.5 & 25/0.5 & 23/0.5 & 25/0.5 & 24/0.5 & \textbf{21/0.4} \\
& Shop Facade~ & 6/0.3 & 6/0.3 & 6/0.3 & 6/0.3 & \textbf{5/0.3} & \textbf{5/0.3} \\
& ~St Mary's Church~ & 17/0.5 & 17/0.5 & 16/0.5 & 17/0.5 & 16/0.5 & \textbf{15/0.5}\\
\midrule
& Average~ & 22/0.4 & 21/0.4 & 20/0.4 & 21/0.4 & 21/0.4 & \textbf{17/0.3} \\
\midrule \morecmidrules
\multirow{6}{*}{\textit{Indoor6}~} & scene1~ & 34.4\% & 34.9\% & 34.9\% & 32.0\% & 30.2\% & \textbf{46.4\%} \\
& scene2a~ & 39.8\% & 40.6\% & 46.1\% & 41.3\% & 42.5\% & \textbf{60.6\%} \\
& scene3~ & 45.2\% & 43.9\% & 44.9\% & 41.3\% & 42.3\% & \textbf{56.4\%} \\
& scene4a~ & 63.2\% & 63.9\% & 62.6\% & 63.2\% & 67.1\% & \textbf{78.7\%} \\
& scene5~ & 19.5\% & 18.1\% & 17.1\% & 18.3\% & 16.9\% & \textbf{22.8\%} \\
& scene6~ & 59.7\% & 60.3\% & 60.0\% & 57.2\% & 58.1\% & \textbf{71.6\%} \\
\midrule
& Average~ & 43.6\% & 43.6\% & 44.3\% & 42.2\% & 42.9\% & \textbf{56.1\%} \\
\bottomrule

\end{tabular}}
\label{table:error_mask}
\end{table}

%% file: tables/ablation.tex
\begin{table}[t]

\centering
\footnotesize

\begin{minipage}[t]{.53\linewidth}
\caption{Ablation study on the impact of each proposed components.}
\resizebox{\linewidth}{!}{
\begin{tabular}{cc|c|c}

\toprule
\textit{Error-Guided } & \textit{Confidence }  & Cambridge & Indoor6  \\ 
\textit{Masks}       & \textit{Refinement}          & (cm/$^\circ$) & (\%)\\
\midrule
\xmark & \xmark & 25/0.4 & 35.6 \\
\cmark & \xmark & 19/0.3 & 41.6 \\
\xmark & \cmark & 19/0.4 & 50.9 \\
\cmark & \cmark & \textbf{17/0.3} & \textbf{56.1} \\
\bottomrule

\end{tabular}}
\label{table:ablation}
\end{minipage}\hfill
\begin{minipage}[t]{.43\linewidth}
\caption{Impact of varying proportions of point prompts ($\tau$).}
\resizebox{\linewidth}{!}{

\begin{tabular}{c|c|c}
\toprule
\textit{Proportions of }  & Cambridge & Indoor6  \\ 
\textit{point prompts ($\tau$)}       & (cm/$^\circ$) & (\%)\\
\midrule
5 & \textbf{17/0.3} & 54.3 \\
10 & \textbf{17/0.3} & \textbf{56.1} \\
15 & \textbf{17/0.3} & 55.4 \\
20 & \textbf{17/0.3} & 55.3 \\
\bottomrule

\end{tabular}}
\label{table:error-prompts}
\end{minipage}

\end{table}

%% file: sections/7_conclusion.tex
\section{Conclusion}
\label{sec::conclusion}
This paper addressed key challenges in SCR for visual localization, and specifically focused on the impacts of dynamic objects and texture-less regions. Our approach introduced an innovative EGFS mechanism through the use of SAM and confidence maps to enhance the performance of SCR. This technique effectively filtered out problematic areas by concentrating on regions with low reprojection errors. Moreover, we used confidence maps to further refine the pixels selected by EGFS and perform this process iteratively to enable dynamic updates of the focused regions. The experimental results on the Cambridge Landmarks and Indoor6 datasets suggest that our method can provide improvements in terms of training efficiency, model size, and accuracy. Our findings highlighted the importance of carefully selecting low-reprojection error pixels by taking into account semantic information and confidence scores. Furthermore, we quantitatively and qualitatively demonstrated that dynamically updating the masks enables more robust selection of error points, thus enabling better training efficiency and effectiveness. Our ablation studies validated the effectiveness of the techniques adopted in our method, and solidified their roles in enhancing the performance.

%% file: sections/acknowledgement.tex
\section*{Acknowledgements}
The authors gratefully acknowledge the support from the National Science and Technology Council (NSTC) in Taiwan under grant numbers MOST 111-2223-E-007-004-MY3, 113-2221-E-007-122-MY3, 113-2640-E-002-003, 113-2221-E-007-104-MY3, as well as the financial support from Woven by Toyota, Inc., Japan. The authors would also like to express their appreciation for the donation of the GPUs from NVIDIA Corporation and NVIDIA AI Technology Center (NVAITC) used in this work. Furthermore, the authors extend their gratitude to the National Center for High-Performance Computing (NCHC) for providing the necessary computational and storage resources.

%% file: main.bbl
\newcommand{\IJCV}{Int. J. Computer Vision (IJCV)}\newcommand{\CVPR}{Proc. IEEE
  Conf. on Computer Vision and Pattern Recognition
  (CVPR)}\newcommand{\CVPRW}{Proc. IEEE Conf. on Computer Vision and Pattern
  Recognition Workshop (CVPRW)}\newcommand{\ICCV}{Proc. IEEE Int. Conf. on
  Computer Vision (ICCV)}\newcommand{\ICCVW}{Proc. IEEE Int. Conf. on Computer
  Vision Workshop (ICCVW)}\newcommand{\ECCV}{Proc. European Conf. on Computer
  Vision (ECCV)}\newcommand{\ECCVW}{Proc. European Conf. on Computer Vision
  Workshop (ECCVW)}\newcommand{\IROS}{Proc. IEEE Int. Conf. on Intelligent
  Robots and Systems (IROS)}\newcommand{\CoRL}{Proc. Conf. on Robot Learning
  (CoRL)}\newcommand{\ICRA}{Proc. IEEE Int. Conf. on Robotics and Automation
  (ICRA)}\newcommand{\AAAI}{Proc. AAAI Conf. on Artificial Intelligence
  (AAAI)}\newcommand{\IJCAI}{Proc. Int. Joint Conf. on Artificial Intelligence
  (IJCAI)}\newcommand{\PAMI}{IEEE Trans. Pattern Analysis and Machine
  Intelligence (TPAMI)}\newcommand{\NIPS}{Proc. Conf. on Neural Information
  Processing Systems (NeurIPS)}\newcommand{\ICML}{Proc. Int. Conf. on Machine
  Learning (ICML)}\newcommand{\ICLR}{Proc. Int. Conf. on Learning
  Representations (ICLR)}\newcommand{\ICLRW}{Proc. Int. Conf. on Learning
  Representations Workshop (ICLRW)}\newcommand{\ICASSP}{Proc. IEEE Int. Conf.
  on Acoustics, Speech, & Signal Processing (ICASSP)}\newcommand{\BMVC}{Proc.
  British Machine Vision Conf. (BMVC)}\newcommand{\ACCV}{Proc. Asian Conf. on
  Computer Vision (ACCV)}\newcommand{\WACV}{Proc. IEEE Winter Conf. on
  Applications of Computer Vision (WACV)}\newcommand{\NAACL}{Proc. Conf. of
  North American Chapter of the Association for Computational Linguistics:
  Human Language Technologies (NAACL)}
\begin{thebibliography}{10}

\bibitem{sam}
Alexander Kirillov, Eric Mintun, Nikhila Ravi, Hanzi Mao, Chloe Rolland, Laura
  Gustafson, Tete Xiao, Spencer Whitehead, Alexander~C. Berg, Wan-Yen Lo, Piotr
  Doll{\'a}r, and Ross Girshick.
\newblock Segment anything.
\newblock {\em arXiv:2304.02643}, 2023.

\bibitem{pnp}
Xiao-Shan Gao, Xiao-Rong Hou, Jianliang Tang, and Hang-Fei Cheng.
\newblock Complete solution classification for the perspective-three-point
  problem.
\newblock {\em IEEE transactions on pattern analysis and machine intelligence},
  25(8):930--943, 2003.

\bibitem{ransac}
Martin~A Fischler and Robert~C Bolles.
\newblock Random sample consensus: a paradigm for model fitting with
  applications to image analysis and automated cartography.
\newblock {\em Communications of the ACM}, 1981.

\bibitem{sarlin2019coarse}
Paul-Edouard Sarlin, Cesar Cadena, Roland Siegwart, and Marcin Dymczyk.
\newblock From coarse to fine: Robust hierarchical localization at large scale.
\newblock In {\em \CVPR}, 2019.

\bibitem{scorf}
Jamie Shotton, Ben Glocker, Christopher Zach, Shahram Izadi, Antonio Criminisi,
  and Andrew Fitzgibbon.
\newblock Scene coordinate regression forests for camera relocalization in
  {RGB-D} images.
\newblock In {\em \CVPR}, 2013.

\bibitem{detone2018superpoint}
Daniel DeTone, Tomasz Malisiewicz, and Andrew Rabinovich.
\newblock Superpoint: Self-supervised interest point detection and description.
\newblock In {\em \CVPR}, pages 224--236, 2018.

\bibitem{sun2021loftr}
Jiaming Sun, Zehong Shen, Yuang Wang, Hujun Bao, and Xiaowei Zhou.
\newblock {LoFTR}: Detector-free local feature matching with transformers.
\newblock In {\em \CVPR}, pages 8922--8931, 2021.

\bibitem{sarlin2020superglue}
Paul-Edouard Sarlin, Daniel DeTone, Tomasz Malisiewicz, and Andrew Rabinovich.
\newblock {SuperGlue}: Learning feature matching with graph neural networks.
\newblock In {\em \CVPR}, 2020.

\bibitem{lindenberger2023lightglue}
Philipp Lindenberger, Paul-Edouard Sarlin, and Marc Pollefeys.
\newblock {LightGlue}: Local feature matching at light speed.
\newblock {\em arXiv preprint arXiv:2306.13643}, 2023.

\bibitem{speciale2019privacy}
Pablo Speciale, Johannes~L Schonberger, Sing~Bing Kang, Sudipta~N Sinha, and
  Marc Pollefeys.
\newblock Privacy preserving image-based localization.
\newblock In {\em \CVPR}, pages 5493--5503, 2019.

\bibitem{anglerepro}
Xiaotian Li, Juha Ylioinas, Jakob Verbeek, and Juho Kannala.
\newblock Scene coordinate regression with angle-based reprojection loss for
  camera relocalization.
\newblock In {\em \ECCVW}, 2018.

\bibitem{fullframe}
Xiaotian Li, Juha Ylioinas, and Juho Kannala.
\newblock Full-frame scene coordinate regression for image-based localization.
\newblock {\em RSS}, 2018.

\bibitem{dsac}
Eric Brachmann, Alexander Krull, Sebastian Nowozin, Jamie Shotton, Frank
  Michel, Stefan Gumhold, and Carsten Rother.
\newblock {DSAC}-{Differentiable RANSAC} for camera localization.
\newblock In {\em \CVPR}, 2017.

\bibitem{dsacpp}
Eric Brachmann and Carsten Rother.
\newblock Learning less is more - {6D} camera localization via {3D} surface
  regression.
\newblock In {\em \CVPR}, 2018.

\bibitem{dsacstar}
Eric Brachmann and Carsten Rother.
\newblock Visual camera re-localization from {RGB} and {RGB-D} images using
  {DSAC}.
\newblock {\em TPAMI}, 2021.

\bibitem{esac}
Eric Brachmann and Carsten Rother.
\newblock Expert sample consensus applied to camera re-localization.
\newblock In {\em \ICCV}, 2019.

\bibitem{ace}
Eric Brachmann, Tommaso Cavallari, and Victor~Adrian Prisacariu.
\newblock Accelerated coordinate encoding: Learning to relocalize in minutes
  using {RGB} and poses.
\newblock In {\em \CVPR}, 2023.

\bibitem{posenet}
Alex Kendall, Matthew Grimes, and Roberto Cipolla.
\newblock {PoseNet}: A convolutional network for real-time {6-DOF} camera
  relocalization.
\newblock In {\em \ICCV}, December 2015.

\bibitem{scenelandmark}
Tien Do, Ondrej Miksik, Joseph DeGol, Hyun~Soo Park, and Sudipta~N. Sinha.
\newblock Learning to detect scene landmarks for camera localization.
\newblock In {\em \CVPR}, 2022.

\bibitem{multioutput}
Abner Guzman-Rivera, Pushmeet Kohli, Ben Glocker, Jamie Shotton, Toby Sharp,
  Andrew Fitzgibbon, and Shahram Izadi.
\newblock Multi-output learning for camera relocalization.
\newblock In {\em \CVPR}, 2014.

\bibitem{forest_rgb}
Eric Brachmann, Frank Michel, Alexander Krull, Michael~Ying Yang, Stefan
  Gumhold, and carsten Rother.
\newblock Uncertainty-driven {6D} pose estimation of objects and scenes from a
  single {RGB} image.
\newblock In {\em \CVPR}, 2016.

\bibitem{forest_rgbd}
Julien Valentin, Matthias Niessner, Jamie Shotton, Andrew Fitzgibbon, Shahram
  Izadi, and Philip H.~S. Torr.
\newblock Exploiting uncertainty in regression forests for accurate camera
  relocalization.
\newblock In {\em \CVPR}, 2015.

\bibitem{ofvl}
Tao Xie, Kun Dai, Siyi Lu, Ke~Wang, Zhiqiang Jiang, Jinghan Gao, Dedong Liu,
  Jie Xu, Lijun Zhao, and Ruifeng Li.
\newblock {OFVL-MS}: Once for visual localization across multiple indoor
  scenes.
\newblock In {\em \ICCV}, 2023.

\bibitem{vsnet}
Zhaoyang Huang, Han Zhou, Yijin Li, Bangbang Yang, Yan Xu, Xiaowei Zhou, Hujun
  Bao, Guofeng Zhang, and Hongsheng Li.
\newblock {VS-Net}: Voting with segmentation for visual localization.
\newblock In {\em \CVPR}, 2021.

\bibitem{sldstar}
Tien Do and Sudipta~N. Sinha.
\newblock Improved scene landmark detection for camera localization.
\newblock In {\em Proceedings of the International Conference on 3D Vision
  (3DV)}, 2024.

\bibitem{pixloc}
Paul-Edouard Sarlin, Ajaykumar Unagar, Måns Larsson, Hugo Germain, Carl Toft,
  Victor Larsson, Marc Pollefeys, Vincent Lepetit, Lars Hammarstrand, Fredrik
  Kahl, and Torsten Sattler.
\newblock {Back to the Feature: Learning Robust Camera Localization from Pixels
  to Pose}.
\newblock In {\em \CVPR}, 2021.

\bibitem{pixselect}
Mohammad Altillawi.
\newblock {PixSelect}: Less but reliable pixels for accurate and efficient
  localization.
\newblock In {\em \ICRA}, pages 4156--4162. IEEE, 2022.

\bibitem{global_instance}
Fei Xue, Ignas Budvytis, Daniel~Olmeda Reino, and Roberto Cipolla.
\newblock Efficient large-scale localization by global instance recognition.
\newblock In {\em \CVPR}, pages 17348--17357, 2022.

\bibitem{sfd2}
Fei Xue, Ignas Budvytis, and Roberto Cipolla.
\newblock {SFD2}: Semantic-guided feature detection and description.
\newblock In {\em \CVPR}, pages 5206--5216, 2023.

\bibitem{d2s}
Bach-Thuan Bui, Dinh-Tuan Tran, and Joo-Ho Lee.
\newblock {D2S}: Representing local descriptors and global scene coordinates
  for camera relocalization.
\newblock {\em arXiv preprint arXiv:2307.15250}, 2023.

\bibitem{FocusTune}
Son~Tung Nguyen, Alejandro Fontan, Michael Milford, and Tobias Fischer.
\newblock {FocusTune}: Tuning visual localization through focus-guided
  sampling.
\newblock In {\em \WACV}, pages 3606--3615, January 2024.

\bibitem{rio10}
Johanna Wald, Torsten Sattler, Stuart Golodetz, Tommaso Cavallari, and Federico
  Tombari.
\newblock Beyond controlled environments: {3d} camera re-localization in
  changing indoor scenes.
\newblock In {\em \ECCV}, pages 467--487, 2020.

\bibitem{dong2021robust}
Siyan Dong, Qingnan Fan, He~Wang, Ji~Shi, Li~Yi, Thomas Funkhouser, Baoquan
  Chen, and Leonidas~J Guibas.
\newblock Robust neural routing through space partitions for camera
  relocalization in dynamic indoor environments.
\newblock In {\em \CVPR}, 2021.

\bibitem{Lee_2021_CVPR}
Donghwan Lee, Soohyun Ryu, Suyong Yeon, Yonghan Lee, Deokhwa Kim, Cheolho Han,
  Yohann Cabon, Philippe Weinzaepfel, Nicolas Guerin, Gabriela Csurka, and
  Martin Humenberger.
\newblock Large-scale localization datasets in crowded indoor spaces.
\newblock In {\em \CVPR}, pages 3227--3236, June 2021.

\bibitem{Fan_2022_CVPR}
Hongyi Fan, Joe Kileel, and Benjamin Kimia.
\newblock On the instability of relative pose estimation and {RANSAC's} role.
\newblock In {\em \CVPR}, pages 8935--8943, June 2022.

\bibitem{vitadapter}
Zhe Chen, Yuchen Duan, Wenhai Wang, Junjun He, Tong Lu, Jifeng Dai, and
  Yu~Qiao.
\newblock Vision transformer adapter for dense predictions.
\newblock {\em arXiv preprint arXiv:2205.08534}, 2022.

\bibitem{efficient_vit}
Han Cai, Junyan Li, Muyan Hu, Chuang Gan, and Song Han.
\newblock {EfficientViT}: Lightweight multi-scale attention for high-resolution
  dense prediction.
\newblock In {\em \ICCV}, pages 17302--17313, 2023.

\bibitem{uncertainty}
Alex Kendall and Yarin Gal.
\newblock What uncertainties do we need in bayesian deep learning for computer
  vision?
\newblock {\em Advances in neural information processing systems}, 30, 2017.

\bibitem{confnet}
Sheng Wan, Tung-Yu Wu, Wing~H Wong, and Chen-Yi Lee.
\newblock Confnet: predict with confidence.
\newblock In {\em 2018 IEEE International Conference on Acoustics, Speech and
  Signal Processing (ICASSP)}, pages 2921--2925. IEEE, 2018.

\bibitem{neumap}
Shitao Tang, Sicong Tang, Andrea Tagliasacchi, Ping Tan, and Yasutaka Furukawa.
\newblock Neumap: Neural coordinate mapping by auto-transdecoder for camera
  localization.
\newblock In {\em Proceedings of the IEEE/CVF Conference on Computer Vision and
  Pattern Recognition (CVPR)}, pages 929--939, June 2023.

\bibitem{sacreg}
{Revaud, J\'er\^ome and Cabon, Yohann and Br\'egier, Romain and Lee, JongMin
  and Weinzaepfel, Philippe}.
\newblock {SACReg: Scene-Agnostic Coordinate Regression for Visual
  Localization}, 2023.

\bibitem{sfm}
Johannes~L. Schonberger and Jan-Michael Frahm.
\newblock Structure-from-motion revisited.
\newblock In {\em \CVPR}, June 2016.

\end{thebibliography}
